%% file: main.tex
\documentclass[conference]{IEEEtran}
\IEEEoverridecommandlockouts

\usepackage{pifont}
\usepackage{colortbl}
\usepackage{tabularx}
\usepackage{cite}
\usepackage{amsmath,amssymb,amsfonts}
\usepackage{algorithmic}
\usepackage{graphicx}
\usepackage{textcomp}
\usepackage{xcolor}
\usepackage{booktabs}
\usepackage{multicol}
\usepackage{multirow}

\begin{document}

\title{A Humanoid Visual-Tactile-Action Dataset for Contact-Rich Manipulation}

\author{%
\IEEEauthorblockN{%
Eunju Kwon\IEEEauthorrefmark{2},
Seungwon Oh,
In-Chang Baek,
Yucheon Park, \\
Gyungbo Kim, 
JaeYoung Moon,
Yunho Choi,
Kyung-Joong Kim\IEEEauthorrefmark{1}\thanks{* Corresponding author: Kyung-Joong Kim (email: kjkim@gist.ac.kr).}}
\IEEEauthorblockA{\textit{Gwangju Institute of Science and Technology (GIST)}\\
Gwangju, South Korea\\ 
ejkwon43@gm.gist.ac.kr}
}

\maketitle

\begin{abstract}
Contact-rich manipulation has become increasingly important in robot learning. However, previous studies on robot learning datasets have focused on rigid objects and underrepresented the diversity of pressure conditions for real-world manipulation. To address this gap, we present a humanoid visual-tactile-action dataset designed for manipulating deformable soft objects. The dataset was collected via teleoperation using a humanoid robot equipped with dexterous hands, capturing multi-modal interactions under varying pressure conditions. This work also motivates future research on models with advanced optimization strategies capable of effectively leveraging the complexity and diversity of tactile signals.

\end{abstract}
 

\section{Introduction}

\input{Sections/1.Introduction}

\section{Related Works}
\input{Sections/2.RelatedWorks}

\section{Visual-Tactile Soft Object Manipulation Dataset}
\input{Sections/3.Method}

\section{Experiments}
\input{Sections/4.Experiments}

\section{Conclusion and Future Work}

\input{Sections/5.Discussion}

\section*{Acknowledgment}

This research was financially supported by the Ministry of Trade, Industry, and Energy (MOTIE), Korea, under the Global Industrial Technology 
Cooperation Center program supervised by the Korea Institute for Advancement of Technology (KIAT) (Grant No. P0028435), and in part by the National Research Foundation of Korea (NRF) grant funded by the Korea government (MSIT) (RS-2025-16902996).

\bibliographystyle{IEEEtran}
\bibliography{ref}

\end{document}

%% file: Sections/1.Introduction.tex
Contact-rich interaction represents a critical gateway for enabling robots to perform complex tasks in real-world environments, yet it remains one of the fundamental challenges in robotic manipulation \cite{cui2021toward}.
Recent advances in multi-modal robotics have demonstrated that incorporating diverse sensing modalities can enhance manipulation performance by providing rich real-world information and expanding the control capabilities of robots \cite{hao2025tla,zhang2025vtla,fang2024rh20t,wan2024vint}.
The fusion of multiple modalities enables robots to perform more sensitive motions and achieve interactive control when dealing with humans or objects with complex shapes and diverse tactile properties.
To tackle the complexity of real-world problems, it is also essential to collect such multi-modal data across diverse robotic embodiments and task domains, ensuring that learned representations generalize beyond a single platform or scenario.

Previous studies on manipulation tasks have primarily focused on rigid objects, without collecting tactile data that accounts for the diverse pressure conditions required during task execution.
In these cases, it is often sufficient to simply detect whether contact occurs \cite{liuvtdexmanip}, which has led many works to adopt relatively simple tactile sensors in order to reduce cost and system complexity \cite{liuvtdexmanip,yang2022touch}.
As a result, much of the existing work remains at a stage that is not yet practical for real-world manipulation. Most prior datasets were collected using simple robotic grippers and surface-level tactile data, which fail to capture the spatially distributed and volumetric nature of real contact interactions, thereby limiting their applicability to contact-rich tasks \cite{hao2025tla,wan2024vint}.
Therefore, to broaden the applicability of robotic research, future datasets must incorporate rich multi-modal signals from humanoid robots, enabling them to better reflect the challenges of real-world manipulation.

To address this gap, we introduce a humanoid-based multi-modal contact-rich dataset.
We collected 101.9K frames of motion data from actual humanoid robots interacting with two soft objects under diverse contact conditions. This paper demonstrates the utility of the dataset through experiments on soft object manipulation tasks and further evaluates it using a state-of-the-art imitation learning baseline to investigate the importance of tactile sensing resolution. Our main contributions are summarized as follows:

{\renewcommand\labelenumi{(\arabic{enumi})}
\begin{itemize}
\item We present, to the best of our knowledge, the first humanoid visual-tactile-action dataset that captures soft objects under diverse control conditions.  

\item We introduce a neural network architecture for the efficient fusion of dense tactile information, enhancing the performance of contact-rich manipulation.  

\item We conduct an in-depth analysis of the dataset, including visualizations that reveal the key characteristics of dense tactile signals.  
\end{itemize}
}

%% file: Sections/2.RelatedWorks.tex
\newcommand{\cmark}{\textcolor{green}{\ding{51}}}
\newcommand{\xmark}{\textcolor{red}{\ding{55}}}

\begin{table*}[ht]
\centering
    \caption{Comparison of Visual-Tactile datasets.}
    \label{tab:datasets}
    \resizebox{\linewidth}{!}{
    \begin{tabular}{l c cccc cccc c c c}
    \toprule
    \multirow{2}{*}{Dataset} & \multirow{2}{*}{Samples} 
    & \multicolumn{4}{c}{Modality} 
    & \multirow{2}{*}{Soft Obj.} & \multirow{2}{*}{Source} 
    & \multirow{2}{*}{Purpose} & \multirow{2}{*}{Resolution} & \multirow{2}{*}{Coverage} \\
    \cmidrule(lr){3-6}
    & & Vision & Language & Touch & Action & & & & & \\
    \midrule
    Touch and Go (2022) \cite{yang2022touch} & 13.9k & \cmark & \xmark & \cmark & \xmark & \cmark & Human & Texture & High & Dense \\
    SSVTP (2023) \cite{kerr2022self} & 4.6k & \cmark & \xmark & \cmark & \xmark & \cmark & Gripper & Texture & High & Dense \\
    PhysiCLeaR (2024) \cite{yuoctopi} & 45.8k & \cmark & \xmark & \cmark & \xmark & \cmark & Human & Pressure & High & Dense \\
    Vint-6D (2024) \cite{wan2024vint} & 2.1M & \cmark & \xmark & \cmark & \cmark & \xmark & Robot(sim+real) & Both & High & Dense \\
    RH20T (2024) \cite{fang2024rh20t} & 50M & \cmark & \xmark & \cmark & \cmark & \xmark & Robot+Human & Both & High & Dense \\
    VTDexManip (2025) \cite{liuvtdexmanip} & 565k & \cmark & \xmark & \cmark & \xmark & \xmark & Human & Pressure & Low & Sparse \\
    TLA (2025) \cite{hao2025tla} & 24k & \cmark & \cmark & \cmark & \xmark & \xmark & Gripper & Both & High & Dense \\
    \rowcolor{gray!15}
    \hline
    \textbf{Ours}       & \textbf{101.9k} & \cmark & \xmark & \cmark & \cmark 
    & \cmark & \textbf{Humanoid} & \textbf{Pressure} & \textbf{High} & \textbf{Dense} \\
    \bottomrule
\end{tabular}}
\end{table*}

\textbf{Visual-Tactile Datasets.}
In this work, we survey visual-tactile datasets as shown in Table~\ref{tab:datasets}. Integration of vision and touch enables robots to infer contact state and physical properties (e.g., texture, stiffness, deformation), remains informative under occlusion and supports pressure-aware control. Early large-scale visual-tactile datasets leveraged image-based high-resolution tactile sensors~\cite{yang2022touch,kerr2022self,yuoctopi,hao2025tla} but largely targeted texture recognition without action labels, limiting direct use for policy learning. Subsequent works added action information~\cite{fang2024rh20t} yet typically lack anthropomorphic hands and humanoid embodiments. Recent efforts on visual-tactile pretraining for anthropomorphic hands and on dense tactile collections~\cite{liuvtdexmanip,wan2024vint} still assume mostly rigid settings and underrepresent soft object. We therefore introduce a humanoid visual-tactile-action dataset explicitly targeting soft objects under controlled pressure for contact-rich manipulation.

\textbf{Contact-rich Manipulation.}
Soft objects are more challenging to manipulate than rigid objects because their shape continuously changes during the manipulation process. Various approaches have been proposed to address this challenge. These include inferring an object’s physical properties from large-scale tactile-language data \cite{yuoctopi}, spatially representing contact by registering visual 3D point clouds and tactile data in a unified coordinate system \cite{huang20253d}, and extracting flow, contact, depth, and force maps from visual-tactile data to construct a unified tactile prior \cite{zhou2024t}. Nevertheless, most existing studies remain limited to simple contact scenarios involving parallel grippers. Contact-rich manipulation of soft objects capable of capturing the complexity of human hand interactions remains an open challenge. 

%% file: Sections/3.Method.tex
\subsection{Data Collection via Humanoid Teleoperation}
\begin{figure}[ht]
  \includegraphics[width=\columnwidth]{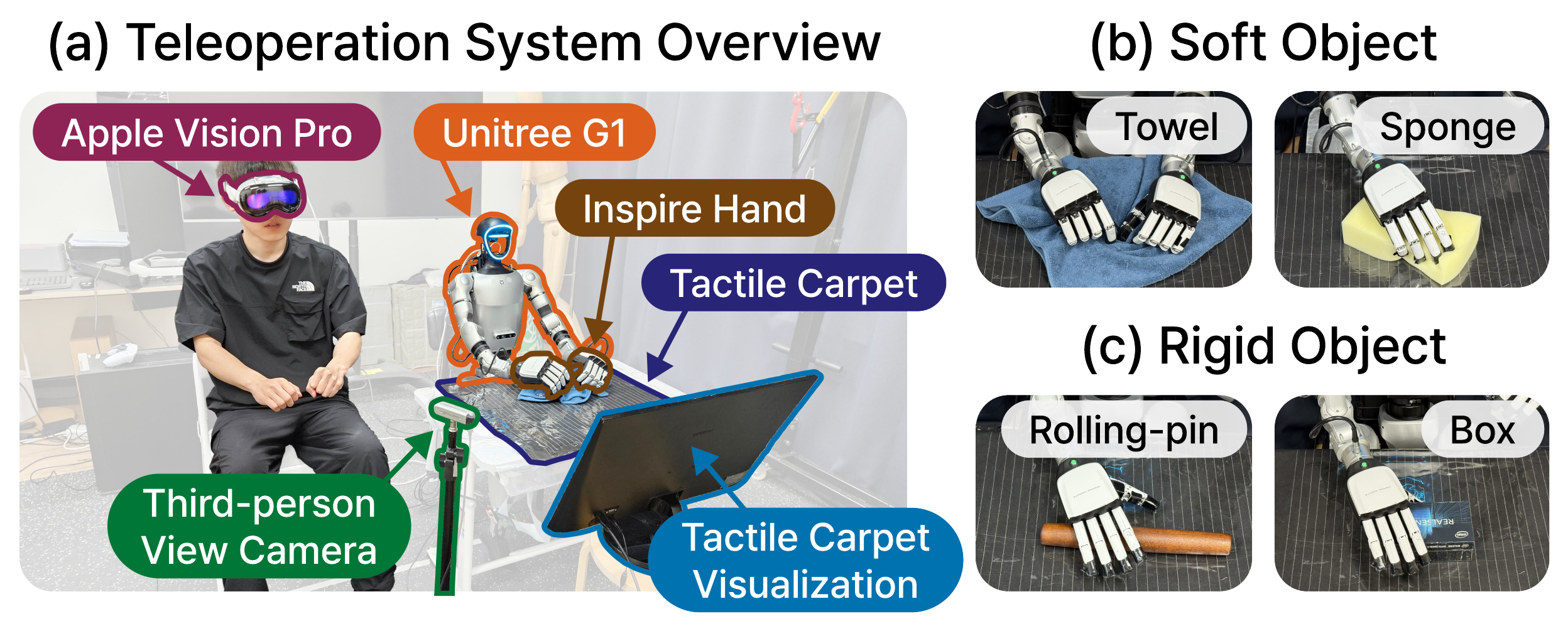}
  \caption{Overview of the teleoperation framework, showing the humanoid robot control setup and the soft and rigid objects used in the experiments.}
  \label{fig:teleop}
\end{figure}

\textbf{Experimental Setup.}
In this section, we describe a humanoid teleoperation setup, as shown in Fig.~\ref{fig:teleop}(a), extending the Unitree workflow to capture rich multi-modal and tactile signals. We record proprioceptive data (arm and finger joint positions), egocentric vision from a head-mounted camera at \(848\times480\), and tactile signals from an Inspire Hand (RH56-DFX, Inspire Robotics) with 1{,}062 high-resolution sensors on each hand, distributed across the fingers and palm. In addition, an Intel RealSense D435 placed 1\,m to the robot’s left provides a third-person view for multi-angle observation. A piezo-resistive tactile carpet~\cite{luo2021intelligent} measures external contact pressure and produces real-time pressure heatmaps.

\textbf{Experimental Objects and Conditions.}
Based on this experimental setup, we collected datasets under different pressure conditions using soft objects to capture pressure variation. To achieve this, we performed tasks using the towel and sponge shown in Fig.~\ref{fig:teleop}(b). For comparison, data were also collected using rigid objects, as shown in Fig. 1(c). The pressure conditions were defined as strong and weak in order to observe how the distribution of tactile signals changes with pressure intensity. Accordingly, four tasks, \emph{Towel Strong}, \emph{Towel Weak}, \emph{Sponge Strong}, and \emph{Sponge Weak} were constructed. During teleoperation, the experimenter executed actions under the predefined pressure conditions, beginning with grasping the object. The tactile carpet heatmap provided real-time feedback on the applied force, ensuring consistent application of the intended pressure condition.

\subsection{Dataset Statistics and Analysis}
\textbf{Dataset Overview and Preprocessing.}
Three experimenters followed the defined protocol to collect the data. Each task consisted of approximately 77–80 episodes (20–30s per episode), resulting in a total of 101.9k samples. The subsequent analysis primarily focused on the tactile signals obtained during this process. All tactile signals were normalized from the raw range of 0–4095 to a scale of 0–1. This normalization was applied to reduce variability in sensor responses and to ensure consistency across different conditions. 

\begin{figure}[t]
  \includegraphics[width=\columnwidth]{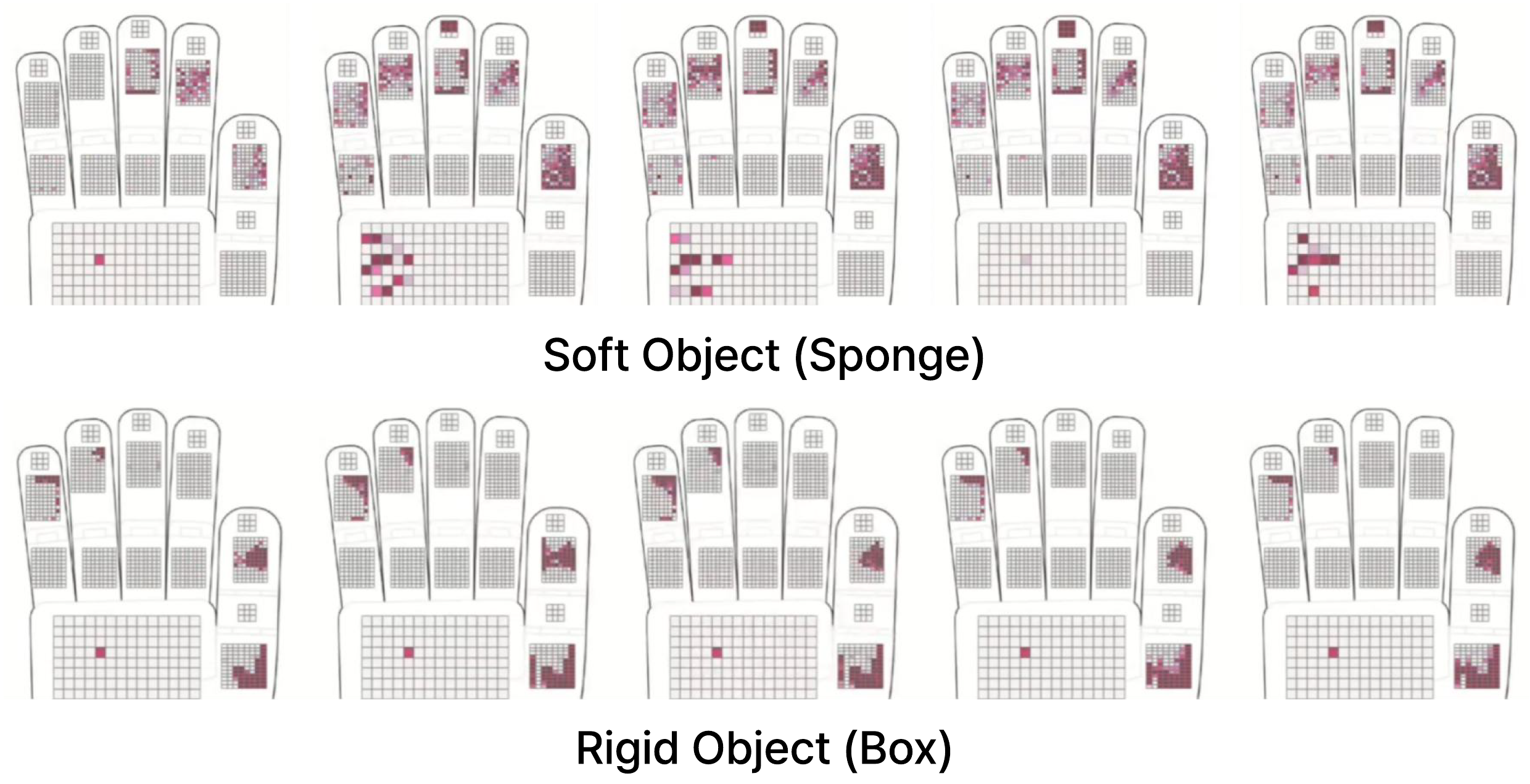}
  \caption{Comparison of tactile signal distributions captured from the dexterous hand when interacting with rigid objects and deformable soft objects, highlighting differences in contact dynamics.}
  \label{fig:results_hand}
\end{figure}

We first analyzed how tactile signals changed over time during the manipulation of rigid and soft objects, as shown in Fig.~\ref{fig:results_hand}. This result highlights the importance of collecting dense tactile signals. The analysis revealed that pressure appeared across diverse contact regions during soft object manipulation, and the distribution evolved dynamically over time. In contrast, for rigid objects, the contact distribution remained constant, exhibiting stable characteristics in both directions and range. These observations indicate that faithfully capturing such time-varying contact patterns requires high-resolution dense tactile sensing.

\begin{figure}[h]
  \includegraphics[width=\columnwidth]{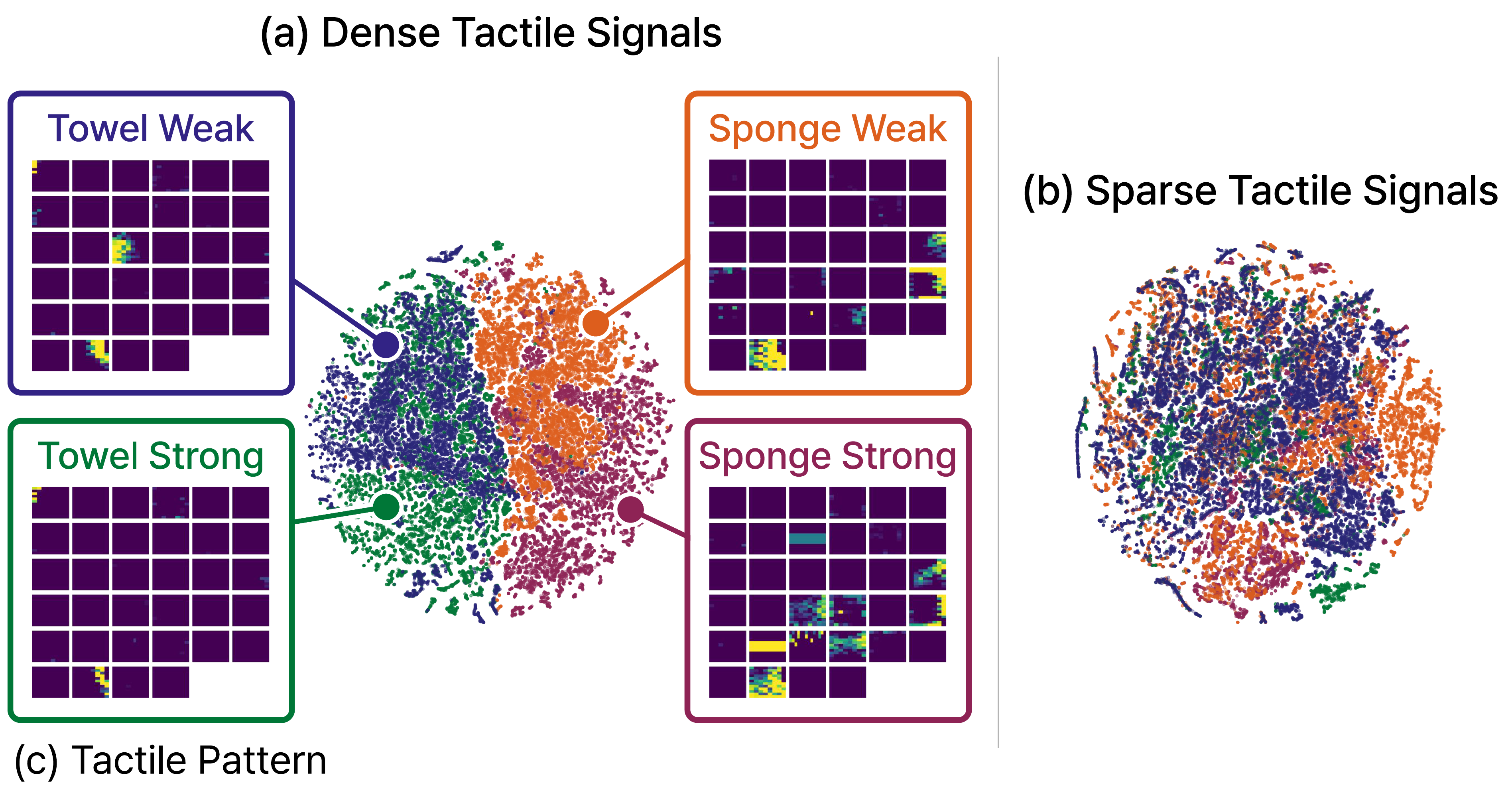}
  \caption{t-SNE embedding of tactile signals under dense and sparse sensing configurations, showing the separability of pressure patterns across different contact conditions.}
  \label{fig:results_t_sne}
\end{figure}

\textbf{Dense vs.\ Sparse Tactile Representations.}
Following the FSR tactile glove proposed in \cite{liuvtdexmanip}, we converted the dense tactile signals into a sparse form to validate the advantages of our high-resolution dense tactile signal collection method. In this conversion, the number of sensors was reduced from 2,124 to 42, representing an approximate 98\% reduction. Fig.~\ref{fig:results_t_sne}(a) visualizes the tactile signals collected across the four tasks. This visualization was performed on the entire dataset to examine the characteristics of the tactile signal samples per task. The t-SNE embedding results clearly separated the distributions of each condition, with samples from the same task forming compact clusters. This demonstrates that our data collection protocol effectively captured tactile information that reflects pressure condition differences. In contrast, the low-resolution sparse tactile signals shown in Fig.~\ref{fig:results_t_sne}(b) did not clearly distinguish differences between tasks. These results suggest that the dense tactile signal reveals task characteristics much more clearly than the sparse representation. 

\textbf{Tactile Pattern under Different Conditions.}
We analyzed which hand regions were primarily utilized for each task during task execution. Fig.~\ref{fig:results_t_sne}(c) visualizes the tactile pattern from each hand patch at a given time step. For simplicity, we divided the hand into patches without further subdivision. Analysis revealed that in the Towel task, relatively fewer tactile signals were observed under strong pressure conditions, whereas in the Sponge task, a greater variety of tactile responses occurred under high pressure conditions. Through these comparative experiments, we confirmed that soft objects induce more changes during manipulation, and that high-resolution, dense tactile signals better classify and understand this information. 

%% file: Sections/4.Experiments.tex
\begin{figure*}[t]
  \includegraphics[width=\textwidth]{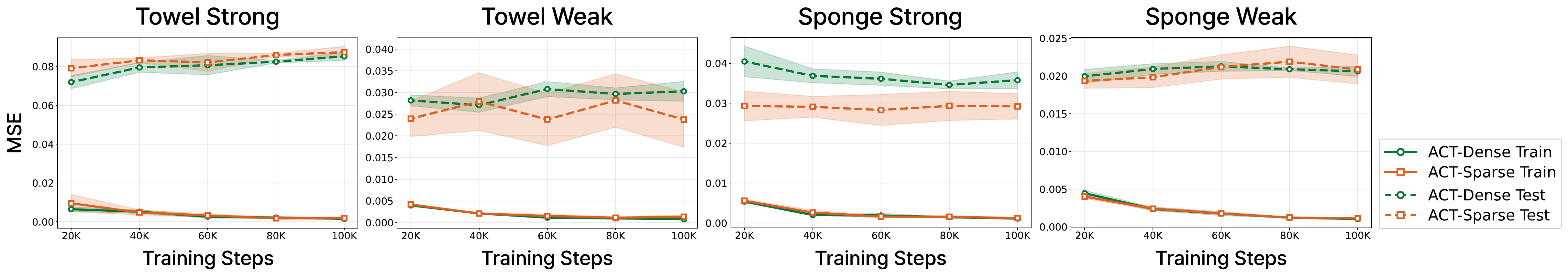}
  \caption{Training and test curves of dense and sparse tactile models across all manipulation tasks.}
  \label{fig:train_results}
\end{figure*}

\subsection{Baseline and Metrics}
In this section, we present experiments designed to verify the effectiveness of the proposed dataset for policy learning under different conditions. To achieve this, we introduced a method for preserving the spatial structure of dense tactile signals. Specifically, the tactile signals obtained from each patch were converted into image representations. Following the approach in \cite{luo2021intelligent}, features were extracted from the 2D tactile map using a CNN, which is a widely adopted method in prior studies. Since the pressure map is represented as a two-dimensional grid structure, it shares the same format as typical image data. Consequently, this representation aligns well with CNN architectures, which are effective at capturing spatial patterns. 

Accordingly, we used these image-based representations as inputs to the model. For training, we employed a state-of-the-art imitation learning baseline Action Chunking Transformer (ACT) \cite{zhao2023learning}. Tactile information was incorporated into the ACT image encoder with visual inputs. At this stage, we compared performance between two input conditions, high-resolution dense tactile signals and low-resolution sparse tactile signals, to quantify the impact of tactile information. The performance of each policy was evaluated using Mean Absolute Error (MAE), which measures the average absolute difference between predicted and ground-truth actions across all timesteps and action dimensions.

\subsection{Experimental Results and Analysis}
\textbf{Training.}
We named the models trained using different tactile datasets ACT-Dense and ACT-Sparse, respectively. These methods were applied to four manipulation tasks Sponge Strong, Sponge Weak, Towel Strong, and Towel Weak. The dataset contained approximately 80 episodes per task, of which 80\% were used for training and 20\% for evaluation. Each model was trained for 100K steps, with evaluations conducted every 20K steps. The training process and performance progression are visualized in Fig.~\ref{fig:train_results}.

\textbf{Results.}
The experimental results are based on training with three different seeds. Across both models, the learning curves consistently exhibited a steady decrease in loss values as training progressed, indicating that the overall differences between the dense and sparse models were not substantial. In contrast, the Towel Weak task exhibited unstable fluctuations in test performance, which may reflect greater uncertainty in tactile signals under weak pressure conditions. For the Sponge tasks, the difference between strong and weak conditions was more pronounced, as the tactile signals varied considerably between the two. This distinction was evident in the test losses, where Sponge Strong and Sponge Weak showed a clear performance gap.

\textbf{Discussion.}
For all tasks, both dense and sparse models demonstrated stable reduction in training loss; however, the test loss did not decrease as significantly, and the performance gap between the two models was relatively small. We attribute this to limitations in the current optimization process. Dense signals, in particular, introduce higher optimization difficulty due to their large dimensionality and high noise levels, thereby requiring more efficient signal processing and optimization strategies  \cite{yao2026flexible}. In conclusion, effectively learning methods for dense tactile signals remains a further challenge. Addressing this issue could be a promising direction to improve robotic intelligence for soft object manipulation.

\subsection{Real-world Deployment}
We also conducted experiments to evaluate the performance of our trained model in real-world environments. For this, the model was tested every 20K steps, beginning from a hand pose that differed from the one used during data collection. During training, the humanoid robot adapted its actions to the specific conditions of each task. In particular, we observed that tasks requiring precise contact control, such as sponge grasping, were more challenging and would benefit from richer data coverage and extended training horizons. A detailed demonstration of our real-world inference process, including representative successful trials, is provided in the supplementary video.

\begin{figure}[ht]
  \includegraphics[width=\columnwidth]{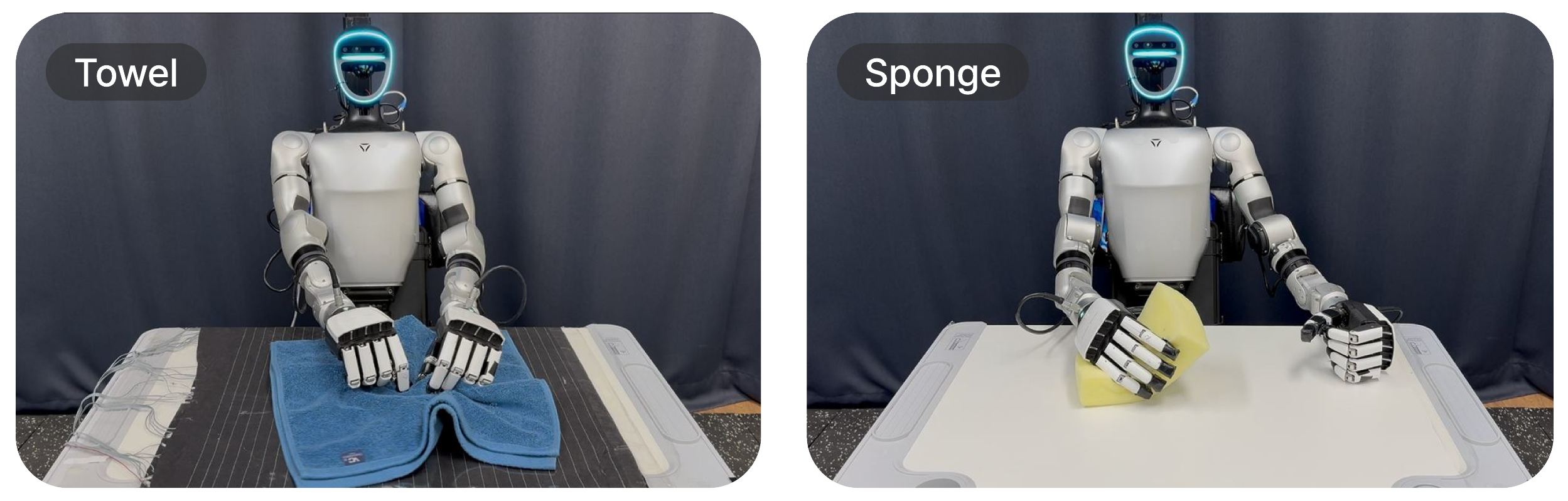}
  \caption{Real-world manipulation experiments.}
  \label{fig:results}
\end{figure}

%% file: Sections/5.Discussion.tex
In this paper, we introduced a humanoid visual-tactile-action dataset capturing tactile responses across diverse pressure conditions when interacting with soft objects. Our analysis showed that soft object manipulation induces high variability in tactile signals, and that high-resolution dense tactile signals are crucial for capturing this complexity. Tasks with more complex tactile patterns tended to show higher loss and greater fluctuations during training. Future work will focus on developing better optimization strategies for learning from complex tactile data, and expanding the dataset with more object types and contact scenarios.